%% file: main.tex
\documentclass[letterpaper]{article} 
\usepackage{aaai25}  
\usepackage{times}  
\usepackage{helvet}  
\usepackage{courier}  
\usepackage[hyphens]{url}  
\usepackage{graphicx} 
\usepackage{natbib}  
\usepackage{caption} 
\usepackage{algorithm}
\usepackage{algorithmic}
\usepackage{amsmath}
\usepackage{xcolor}

\newcommand{\ignorethis}[1]{}
\usepackage{multirow}

\usepackage{newfloat}
\usepackage{listings}
\DeclareCaptionStyle{ruled}{labelfont=normalfont,labelsep=colon,strut=off} 
\floatstyle{ruled}
\newfloat{listing}{tb}{lst}{}
\floatname{listing}{Listing}

\usepackage{amsmath}

\title{DriveGazen: Event-Based Driving Status Recognition using Conventional Camera}

\author{
    Xiaoyin Yang\\
    Dalian University of Technology\\
    tooyoungalex@outlook.com
}

\usepackage{bibentry}

\begin{document}

\maketitle

\begin{abstract}
We introduce a wearable driving status recognition device and our open-source dataset, along with a new real-time method robust to changes in lighting conditions for identifying driving status from eye observations of drivers. The core of our method is generating event frames from conventional intensity frames, and the other is a newly designed Attention Driving State Network (ADSN). Compared to event cameras, conventional cameras offer complete information and lower hardware costs, enabling captured frames to encode rich spatial information. However, these textures lack temporal information, posing challenges in effectively identifying driving status. DriveGazen addresses this issue from three perspectives. First, we utilize video frames to generate realistic synthetic dynamic vision sensor (DVS) events.Second, we adopt a spiking neural network to decode pertinent temporal information. Lastly, ADSN extracts crucial spatial cues from corresponding intensity frames and conveys spatial attention to convolutional spiking layers during both training and inference through a novel guide attention module to guide the feature learning and feature enhancement of the event frame. We specifically collected the Driving Status (DriveGaze) dataset to demonstrate the effectiveness of our approach. Additionally, we validate the superiority of the DriveGazen on the Single-eye Event-based Emotion (SEE) dataset. To the best of our knowledge, our method is the first to utilize guide attention spiking neural networks and eye-based event frames generated from conventional cameras for driving status recognition.Please refer to our project page for more details: https://github.com/TooyoungALEX/AAAI25-DriveGazen.
\end{abstract}

%
\input{sections/introduction}

\input{sections/related_work}

\input{sections/drivegazen}

\input{sections/dataset}

\input{sections/assessment}
\input{sections/conclution}

\clearpage
\bibliography{aaai25}
\clearpage

\end{document}

%% file: sections/introduction.tex
\section{Introduction}

Driver state and behavior are crucial to traffic safety. Factors such as the driver's attention level, driving condition, and fatigue directly impact their perception and response to road situations. Developing effective technologies to identify and monitor driver states has become a significant research direction in traffic safety. However, predicting driver states from conventional RGB images is a challenging task; spatial and temporal cues from driving conditions can be adversely affected by head posture and partial occlusion. Existing facial recognition models for classifying driving states in RGB frames are built on complex CNN-based models, such as ResNet 50, Transformer, and Inception-based methods. Different lighting conditions and fast user movements make driver state recognition more complicated, and despite cumbersome large network enhancement modules, driver state recognition from RGB images remains difficult and fragile.
We will introduce a novel wearable driver state recognition prototype where users only need to wear a pair of glasses (DG3). Mobile wearable devices can provide stronger feature capture under rapid head movements while offering high resolution for capturing more spatial features and higher temporal resolution for capturing more temporal features. Even though this device provides a stable fixed view of both eyes and traditional camera technology is mature and low-cost, estimating driver states from eye features still faces unique challenges.A key issue is that traditional cameras cannot effectively resolve temporal information under limited lighting conditions. These temporal features are not only crucial for driver state recognition but also important for inferring more informative spatial features. For example, while the eye sockets are major spatial cues, they provide less information for driving state classification. In contrast, subtle movements related to facial units, such as raising the outer eyebrows and squinting, provide stronger cues for eye-based driving state recognition.
To address these challenges, we designed the DriveGazen method, which first generates realistic synthetic dynamic visual sensor (DVS) events from video frames and employs the Attention Driving State Network (ADSN) to combine the best features of events and intensity frames, guiding asynchronous event-based driver state recognition with spatial texture cues from the corresponding intensity frames.To train our lightweight eye-based driving state network (ADSN) and stimulate research on event-based eye driver state recognition, we collected a new eye-based event driving state (DriveGaze) dataset. We validated our method on the DriveGaze dataset and demonstrated state-of-the-art driver state recognition capabilities, achieving a significant improvement of 3\% in both WAR and UAR compared to the second-best method.

Specifically, our work makes the following five contributions: 
\begin{itemize}
	\item A novel real-time driver state recognition method based on low-cost conventional camera;
	\item Utilizing video frames to generate realistic synthetic dynamic vision sensor (DVS) events;
	\item A low-latency spiking neural network with guide attention suited for in-the-wild deployment;
    \item The first publicly available eye-based event-driven driving state dataset generated from conventional cameras, containing intensity frames and corresponding events, capturing data from different ages, races, genders, etc;
    \item Validating the superiority of the DriveGazen method on the Single-eye Event-based Emotion (SEE) dataset.
\end{itemize}

\textit{Limitations.}
DriveGazen partially relies on events generated from intensity frames, which may lead to performance degradation when variations are minimal. While our method effectively handles most scenarios, as confirmed by our experimental results, further improving robustness is an exciting direction for future research in eye-based driving state recognition. 


%% file: sections/related_work.tex
\section{Related Work}
\begin{figure*}[ht]
  \centering
  \includegraphics[width=\textwidth]{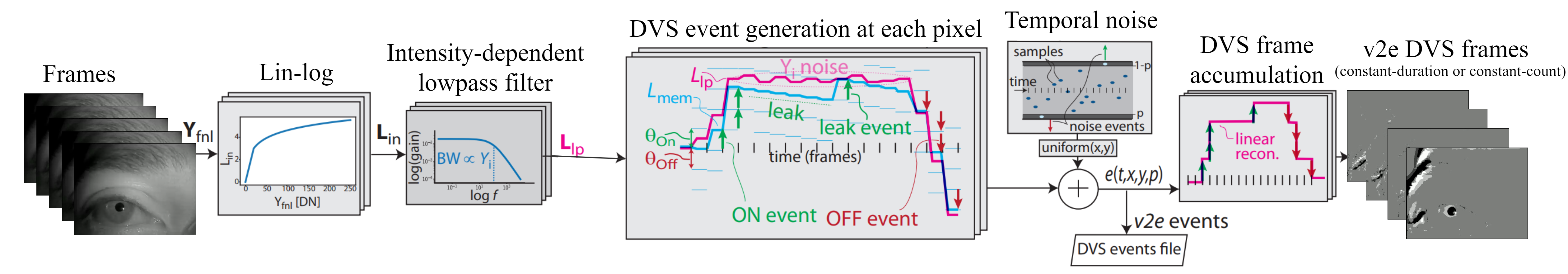}
  \caption{Steps of the v2e DVS event generation (adapted from \cite{hu2021v2e})}
  \label{fig:dvs}
\end{figure*}

We focus on measuring driving status and recognition.Then, we explain the spiking neural network mainly used.

\subsubsection{Driving Status Sensing Methods.}
Researchers utilize physiological measurements such as electroencephalography (EEG), functional near-infrared spectroscopy (fNIRS), electrocardiography (ECG), electrodermal activity (EDA), and respiration (RESP) to identify driver states\cite{wan2019driving}. For example, changes in states can lead to variations in facial temperature, hence there are studies employing facial infrared thermal imaging techniques for identification purposes\cite{zhang2019discriminating}. Additionally, some research examines driver states by collecting their hormone levels\cite{taamneh2017multimodal}.
Apart from physiological measurements, behavioral measurements are also employed to gauge driver states. Some studies utilize near-infrared (NIR) facial expression recognition methods to identify driver states more accurately\cite{gao2014detecting}. Simultaneously, by analyzing dialogues between drivers and in-vehicle information systems, researchers have found that using voice recognition can identify various states\cite{jones2008using}. Moreover, there are studies that identify driver states through posture movements, validating the feasibility of using radio frequency (RF) technology for state recognition\cite{raja2018towards}.
In addition to physiological and behavioral measurements, some studies propose collecting information about driver, vehicle status, and changes in the surrounding environment to infer the driver's state\cite{wan2019driving}. For instance, based on the pressure characteristics of the throttle and brake pedals, classify the driver's happy and unhappy states\cite{nor2010driver}. Furthermore, utilizing inertial measurement units (IMUs) to detect driver states is also a common method\cite{lee2017wearable}. Additionally, some studies use self-report scales to measure driver states, such as Positive and Negative Affect Schedule (PANAS), Self-Assessment Manikin (SAM), and Differential Emotions Scale (DES)\cite{jeon2011angry}.

\subsubsection{Driving Status Recognition Algorithms.}
Researchers typically employ supervised machine learning for implementation. Lee et al.\cite{lee2017wearable} successfully classified three driving states based on PPG, EMG, and IMU signals using SVM, achieving an accuracy of $99.52$\%. Ooi et al.\cite{ooi2016driver} and Gao et al.\cite{gao2014detecting} utilized SVM to classify driver states based on EDA and FEA signals, achieving an accuracy of $85$\%. Other SVM-based development algorithms are also frequently used for state recognition. For example, Wan et al.\cite{wan2019driving} used Least Squares Support Vector Machine (LS-SVM) to detect states based on multimodal signals. Another commonly used algorithm is k-Nearest Neighbors (kNN); Raja et al. employed this method\cite{raja2018towards} to classify anger and neutral states. Nor and Wahab\cite{nor2010driver} used Multi-Layer Perceptron (MLP) to recognize driver states based on velocity and accelerator pedal position. Other traditional machine learning algorithms (such as Bayesian networks) are also used for state recognition\cite{rebolledo2014developing}. Deep learning algorithms have also been successfully implemented in driver state recognition. Lee et al.\cite{lee2018convolutional} collected near-infrared and thermal image data of driver's faces and used Convolutional Neural Networks (CNN) to classify driver's anger and neutral states, achieving a recognition accuracy of $99.96$\%. Although detection accuracy in various studies sometimes reaches $99.96$\%, most studies are conducted on different datasets. Different recognition tasks, data collection methods, and even different expressions of the same state category can all affect recognition accuracy in driver state detection.

\subsubsection{Spiking Neural Network}
Unlike artificial neural networks (ANN) that are purely digitally coded and whose input and output are numerical values, spiking neural networks (SNN) simulate biological processes, include the concept of time, and only exchange information (pulse), with input and output being pulse sequences. SNN describe the properties of units in the nervous system with varying degrees of detail. SNN simulate three states of biological neurons: resting, depolarized, and hyperpolarized\cite{ding2022biologically}. When a neuron is in a resting state, its membrane potential remains constant and is usually set to 0. An increase in membrane potential is called depolarization; conversely, a decrease in membrane potential is hyperpolarization. When the membrane potential is above the potential threshold, an action potential, or pulse, is triggered, and a binary-valued pulse signal is used as output to transmit information between neurons. SNN are low-energy biomimetic methods that work in continuous time using discrete signals such as pulses. They can accept the sparsity found in biology and are compatible with high temporal resolution. SNN balance accuracy and computational feasibility. Existing facial driver state recognition methods can usually only identify the peak emotional state or a single driving state in the entire sequence, and are therefore not suitable for applications that require robust estimation of intermediate states. We introduce a lightweight guide attention driving state mothod (DriveGazen) that can effectively recognize various states using SNN. DriveGazen not only remembers the peak phases of individual driving states, but also exploits temporal cues to distinguish different phases, using frames captured by traditional cameras as input and generating event frames. We adopt a hybrid system that utilizes spatial cues and traditional intensity frames to guide temporal feature extraction during training and inference.

%% file: sections/drivegazen.tex
\section{DriveGazen}
\label{sec:DriveGazen}

The DriveGazen method first generates realistic synthetic dynamic visual sensor (DVS) events from video frames and utilizes the Attention Driving State Network (ADSN) to combine the best features of events and intensity frames, guiding asynchronous event-based driver state recognition with spatial texture cues from the corresponding intensity frames. Next, I will provide a detailed description of each part of the method.

\subsection{Video to Event(v2e)}

We convert RGB video into grayscale frames, where pixel values are treated as luma intensity values. Figure \ref{fig:dvs} illustrates the process of synthetic event generation for a single DVS pixel\cite{hu2021v2e}. \( Y_{fnl} \) represents the sampled frame. We denote \( Y \) as the pixel’s luma intensity value in a luma frame \( Y \). Similarly, \( L \) represents the pixel’s log intensity values in a log intensity frame \( L \).

Standard digital video represents intensity linearly, while DVS pixels detect changes in log intensity. For luma intensity values \( Y < 20 \) digital numbers (DN), we use a linear mapping from exposure value (intensity) to log intensity to reduce quantization noise in the synthetic DVS output.

Since real DVS pixels have finite analog bandwidth, an optional low-pass filter is used to filter the input \( L \) value. The v2e model simulates this effect by making the filter bandwidth (BW) increase monotonically with the intensity value. Although the photoreceptor and source follower form a 2nd-order low-pass filter, one pole usually dominates, so the filter is implemented as an infinite impulse response (IIR) first-order low-pass filter. The nominal cutoff frequency is \( f_{3dB}{max} \) for full white pixels. The filter’s bandwidth is proportional to the luma intensity values \( Y \). We denote the filtered \( L \) value as \( L_{lp} \). The shape of the filter’s transfer function is shown in Figure \ref{fig:dvs}. To avoid nearly zero bandwidth for small DN pixels, an additive constant limits the minimum bandwidth to about 10\% of the maximum value.

We assume the pixel has a memorized brightness value \( L_{mem} \) in log intensity, and the new low-pass filtered brightness value is \( L_{lp} \). The model then generates a signed integer quantity \( N_e \) of positive ON or negative OFF events from the change \( \Delta L = L_{lp} - L_{mem} \), where \( N_e = \lfloor \Delta L / \theta \rfloor \). If \( \Delta L \) is a multiple of the ON and OFF thresholds, multiple DVS events are generated. The memorized brightness value is updated by \( N_e \) multiples of the threshold.

DVS pixels emit spontaneous ON events called leak events\cite{nozaki2017temperature}, with a typical rate of approximately 0.1 Hz. These events are caused by junction leakage and parasitic photocurrent in the change detector reset switch\cite{nozaki2017temperature}. The v2e model adds these leak events by continuously decreasing the memorized brightness value \( L_{mem} \). The leak rate varies according to random fluctuations in the event threshold, decorrelating leak events across different pixels.

The quantal nature of photons leads to shot noise: if, on average, \( K \) photons are accumulated in each integration period, then the average variance will also be \( K \). At low light intensities, the effect of shot noise on DVS output events increases significantly, resulting in balanced ON and OFF shot noise events at rates above 1 Hz per pixel. The v2e model simulates temporal noise using a Poisson process. It generates ON and OFF temporal noise events to match a noise event rate \( R_n \) (default 1 Hz). To model the increase in temporal noise with reduced intensity, the noise rate \( R_n \) is multiplied by a linear function of luma \( 0 < Y \leq 1 \), which reduces noise in brighter areas by a factor \( 0 < c < 1 \) (default \( c = 0.25 \)). This modified rate \( r \) is multiplied by the time step \( \Delta t \) to obtain the probability \( p = r \times \Delta t \leq 1 \), which is applied to the next sample. For each sample, a uniformly distributed number in the range 0-1 is compared against two thresholds \([p, 1-p]\) to determine if an ON or OFF noise event is generated. These noise events are added to the output and reset the pixels.Please refer to the supplementary materials for more details.

\subsection{Attention Driving State Network(ADSN)}

\begin{figure}[h!]
  \centering
  \includegraphics[width=\linewidth]{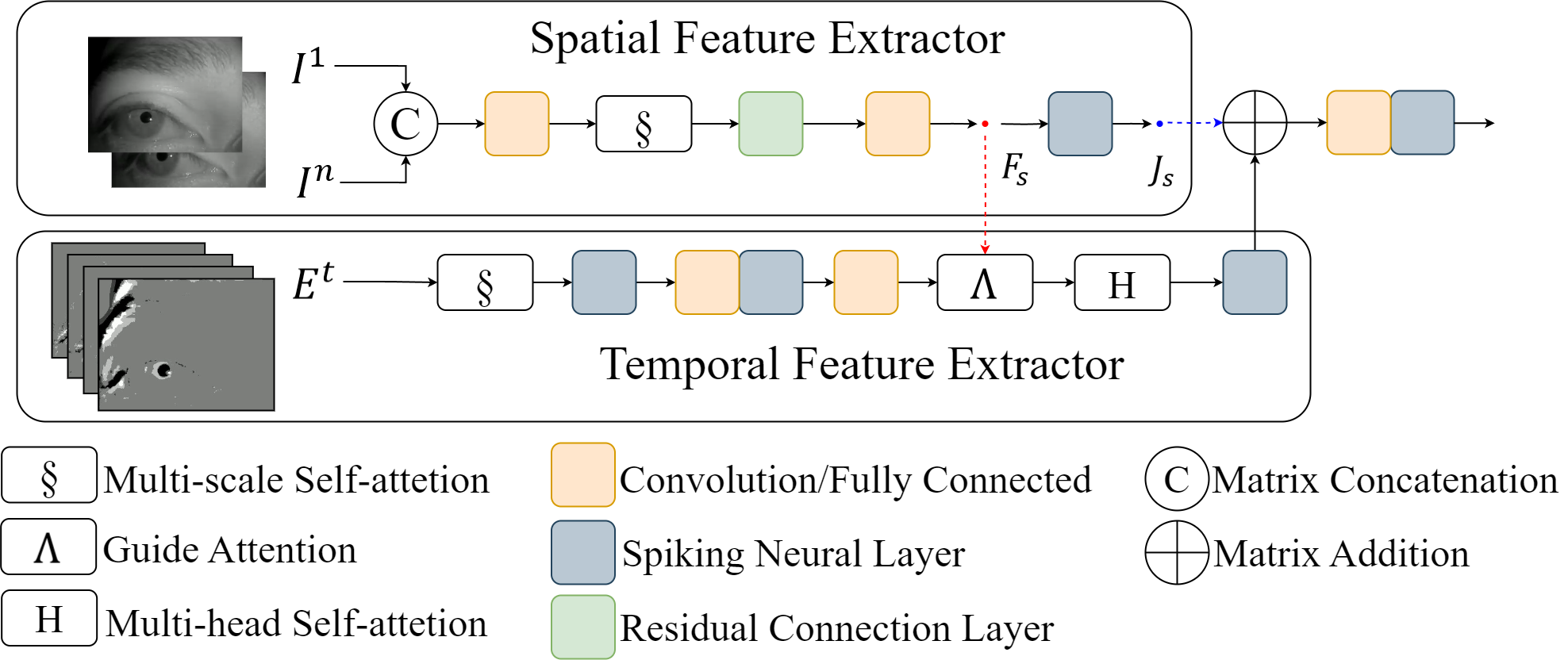}
  \caption{Attention Driving State Network}
  \label{fig:DriveGazen_pipeline}
\end{figure}

As illustrated in Figure \ref{fig:DriveGazen_pipeline}, Specifically, ADSN includes spatial and temporal feature extractors and a guiding attention module. 
The spatial feature extractor achieves spatial feature extraction by decoupling the sequence length, extracting spatial information only from the first and last frames of the grayscale sequence. ADSN aggregates asynchronous events captured between two grayscale frames into \( n \) synchronized event frames. The core of the temporal feature extractor is the spiking neural layer (SNN), which makes decisions based on membrane potential to remember temporal information from previous event frames. Unlike RNN\cite{nah2019recurrent,kag2021time}, SNN can learn temporal dependencies of arbitrary length without special handling. 
The guide attention module uses spatial cues from the spatial feature extractor to guide feature learning and enhancement of event frames. It also uses the spiking neural layer to transform spatial features \( F_s \) into spikes \( J_s \), which are then fused with the event frames. Next, I will provide a detailed description of each module of the network.

\subsubsection{Spatial Feature Extractor $S$}
\label{sec:S}
As illustrated in Figure\ref{fig:DriveGazen_pipeline}, to decouple spatial feature extraction from grayscale sequences in terms of sequence length, the extractor only extracts spatial information from the first frame $I_1$ and the last frame $I_n$ of the sequence. This not only reduces the reliance on grayscale frames, but also improves recognition performance compared to using all frames, as shown in experiments.
First, the two grayscale frames are concatenated in the channel dimension, and then the channel dimension is restored to the original number of channels through a $1 \times 1$ convolution kernel, in order to keep consistent with the convolution operation of the temporal feature extractor(Equation \ref{eq:l_s}).
Next, multi-scale spatial features are extracted and fused through a multi-scale self-attention module. This allows the network to learn small-scale action unit information and also consider the joint information of larger-scale action units while using residuals to reduce information loss(Equation \ref{eq:S_fs}).
Then, two $3 \times 3$ convolution kernels further extract high-level features in the spatial dimension. In order to better retain the temporal features extracted by the temporal feature extractor, we also use a convolution-spiking neural layer to convert the spatial features $F_s$ into a pulse form $J_s$(Equation \ref{eq:Fs_Js}), and add it to the temporal features to enhance feature discrimination.
 Formally, the spatial feature extractor can be defined as:
\begin{align}
    J_s & = \Phi^1(F_s)\label{eq:Fs_Js}\\
    F_s &= C_3(C_3(\S_{(3,5,7)}(l_s)+l_s))\label{eq:S_fs} \\
    \S(i_1, \dots, i_n)(\cdot) & := C_1\left(\left[\omega_1^s C_{i_1}(\cdot), \dots, \omega_n^s C_{i_n}(\cdot)\right]\right)\label{eq:S_Omega}\\
    \omega_1^s, \dots, \omega_n^s &= \sigma\left([\Upsilon(C_{i_1}(l_s)), \dots, \Upsilon(C_{i_n}(l_s))]\right)\label{eq:omega}\\
    \Upsilon(\cdot) &:= C_1(\mathcal\varphi(C_1(\mathcal{A}(\cdot))))\label{eq:Upsilon}\\
    l_s &= C_1([I^1, I^n])\label{eq:l_s}
\end{align}
where $(\cdot)$ denotes channel-wise concatenation; $C_i$ and $\sigma$ denote an $i \times i$ convolutional layer and a softmax function, respectively; $i_1, \dots, i_n$ denote the value of $i$ in the convolutional layer $C_i$ of the multi-scale self-attention module. According to the equation \ref{eq:S_fs}, the values here are 3, 5, 7 respectively. $\mathcal{A}$ denotes the adaptive average pooling layer; $\varphi$ is a serial operation of a batch normalization operation and a ReLU activation function; $\Phi^t$ is a spiking layer that keeps membrane potential from the previous time step, $t-1$. The initial membrane potential, $t = 0$ (see Equation \ref{eq:SNN_layer}).Which only realizes the conversion from floating-point features to 0-1 pulse features without maintaining any temporal information.

\subsubsection{Temporal Feature Extractor $T$}
\label{sec:T}
The core architecture of the temporal feature extractor is a spiking neural network. Spiking neurons output spike signals based on the accumulation, decay, and reset mechanism of membrane potential to capture the temporal trend in the input sequence. When the membrane potential exceeds a threshold, an action potential (i.e., a spike) is triggered and the membrane potential is reset. The triggering process itself is non-differentiable and cannot be trained by traditional stochastic gradient descent optimization methods. Instead, this paper adopts spatio-temporal backpropagation (STBP) and a convolution-spiking neural layer \cite{wu2018spatio} to circumvent this problem. The convolution-spiking neural layer uses a convolution-based layer for signal aggregation and a LIF-based spiking neural layer \cite{gerstner2002spiking} to manage the potential decay and reset process. This modification makes it possible to learn different accumulation strategies by leveraging convolution-based methods and allows spiking neurons to operate effectively in the temporal domain.
The temporal feature extractor receives a total of $n$ event frames as input, denoted as $E^1$ to $E^n$, and processes each frame in chronological order. Figure \ref{fig:DriveGazen_pipeline} illustrates the architecture of the temporal feature extractor. Formally, after receiving the spatial features $F_s$ and pulse features $J_s$ from the spatial feature extractor, the temporal feature extraction process of $E^t$ can be expressed by equations \ref{eq:T_O} to \ref{eq:SNN_layer}:

\begin{align}
 O^{t} &= M\left(\tau\left(\tau\left(J_{c}^{t}\right)\right)\right) \label{eq:T_O}\\
 J_{c}^{t} &= \Phi^t(J_{e}^{t}) \oplus J_s\\
 J_{e}^{t} &= H\left(\Lambda(F_s, F_{e}^{t})\right)\\
 F_{e}^{t} &= C_3(\Phi^t(C_3(\Phi^t(\S'_{(3,5,7)}(E^t))))) \label{eq:T_fe}\\
 \tau(\cdot) & := \Phi^t\left(\Psi(\cdot)\right)\\
 \S'\left(i_1, \dots, i_n\right)(\cdot) & := C_1\left(\left[\omega_1^s C_{i_1}(\cdot), \dots, \omega_n^s C_{i_n}(\cdot)\right]\right)
\end{align}

$\S'$ represents the same structure as the multi-scale self-attention module $\S$ in the spatial feature extractor. $\Psi$ is a fully connected layer; $M$ extracts the membrane potential from the spiking neural layer, and $\Lambda$ is the guide attention module. $H$ is multi-head self-attention. $\Phi^t(\cdot)$ is a spiking neural layer that records the previous spiking state $P^{t-1}$ and accumulated membrane potential $V^{t-1}$. Upon receiving a new input stimulus $X^t$, the membrane potential adjusts based on the previous pulse emission and accumulates the new stimulus. The spiking neural layer emits updated pulses $P^t$ and updates the membrane potential $V^t$ as follows:
\begin{align}
    P^t & =h(V^t-\Theta)\\ 
    V^t & =\alpha V^{t-1}(1-P^{t-1}) + X^t\\
    h(x) &= \begin{cases}0 & x<0 \label{eq:SNN_layer}\\
    1 & x>=0 \end{cases}
\end{align}
$\Theta$ is the membrane potential threshold, set to 0.3 in this experiment. The parameter $\alpha$ is the attenuation factor for hyperpolarization. The potential $V^t$ is updated such that, for a spike at $t-1$, the membrane potential resets to $0$ by scaling $1-P^{t-1}$, with $X^t$ as the corresponding input. Finally, the driving state is predicted based on the average value of $O_t$ for $t \in [1, n]$, as defined in equation \ref{eq:final_result}:

\begin{align}
    R &= \sigma(\frac{1}{n}\sum_{t=1}^{n} O^{t})
\label{eq:final_result}
\end{align}
where $\sigma$ is a Softmax activation function.

\begin{figure}[h!]
  \centering
  \includegraphics[width=\linewidth]{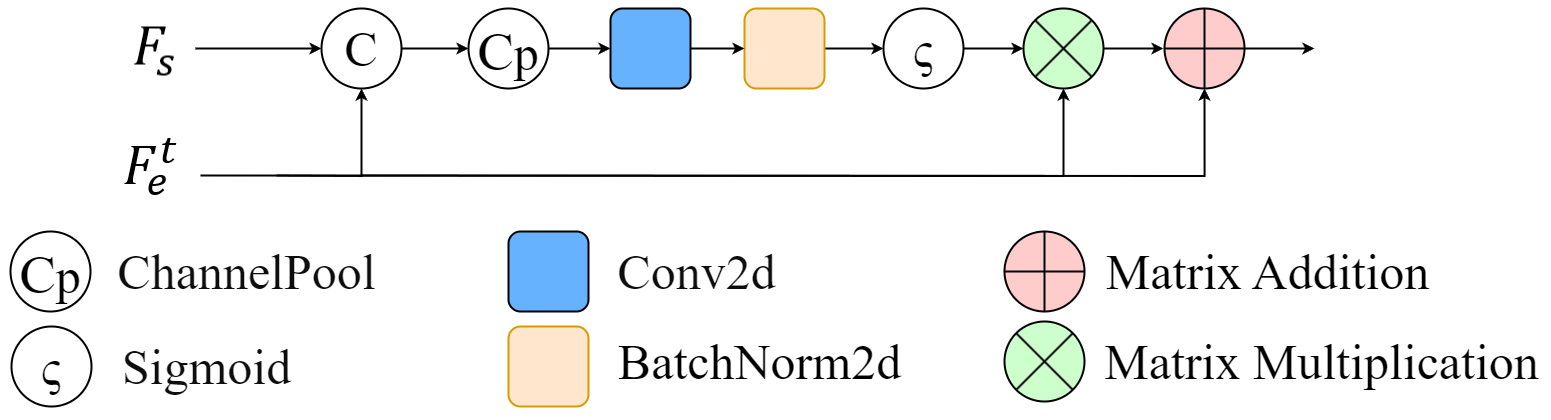}
  \caption{Guide Attention Module}
  \label{fig:GuideAttention}
\end{figure}
\subsubsection{Guide Attention Module $\Lambda$}
\label{sec:Λ}
Due to the lack of reliable texture information in the event domain, relying solely on event information cannot generate an effective solution. Therefore, we utilize spatial features extracted from grayscale frames to inject rich texture clues into the temporal feature extractor. To guide grayscale frames to guide event frames from the spatial domain to the temporal domain, firstly, the spatial cues \(F_S\) learned by the spatial feature extractor are passed through a channel attention module. The learned temporal attention mechanism scores are then allocated to the event frames to enhance the temporal clues. Further spatial attention learning is conducted on the enhanced grayscale frames, and then the spatial attention scores are allocated to the event frames. Finally, the strengthened temporal clues are obtained by adding them to the event frames. The operation process is defined as equations \ref{eq:F_cp} to \ref{eq:GA}, and the design diagram of the guided attention module is shown in Figure \ref{fig:GuideAttention}.
\begin{align}
    F_{cp} &= F^t_e \times F_{p} + F^t_e \label{eq:F_cp}\\ 
    F_{p} &=  \zeta(\mathcal\varphi(C_1(F_{c})))\\
    F_{c} &= C_P(\Delta)\\
    \quad \Delta &= (F_{S})\cdot (F^t_e) \label{eq:GA}
\end{align}

%% file: sections/dataset.tex
\begin{figure}[ht!]
  \centering
  \begin{tabular}{c}
\includegraphics[width=0.95\linewidth]{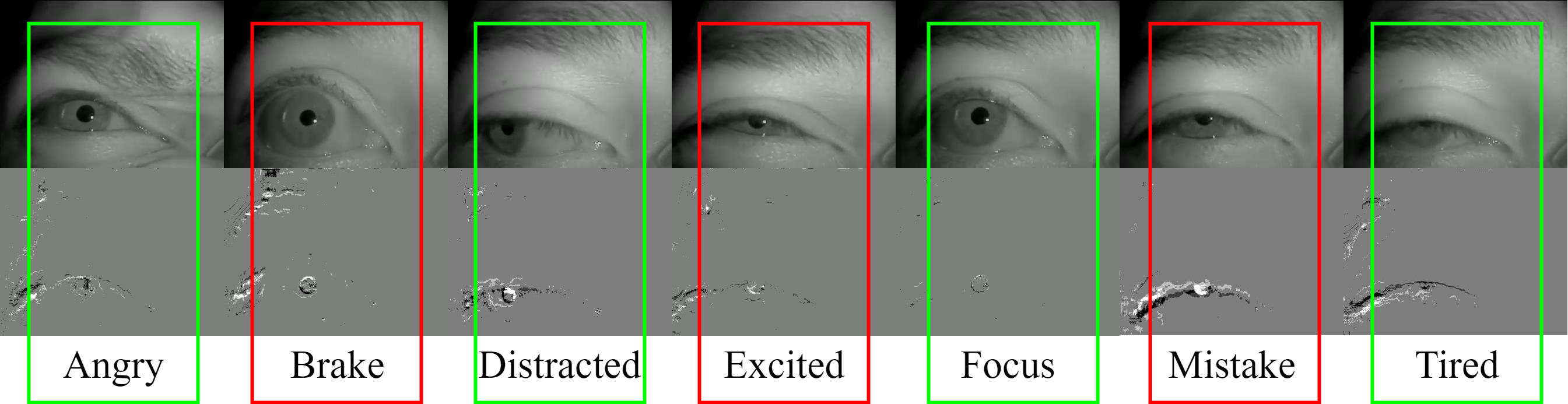} 
 \end{tabular}
  \caption{The newly collected  Event-based Driving Status (DriveGaze) dataset covers seven classes.}
  \label{fig:sta_dataset}
\end{figure}

\section{Dataset}
To the best of our knowledge, there currently does not exist a dataset for driving state recognition based on eye events captured by conventional cameras. To address the lack of training data for event-based driving state recognition, we collected a new event-based driving state dataset (DriveGaze); see Figure \ref{fig:sta_dataset}.  DriveGaze consists of driving state eye data from $47$ volunteers of different ages, genders, and races, captured using the DG3 eye tracker's conventional camera and converted into event frames; see Figure \ref{fig:dvs}. The DG3 camera is positioned in front of both eyes, with a resolution of $384$*$288$ and a frame rate of 60FPS. Unlike the previously mentioned datasets, our wearable device avoids screen occlusion, resulting in clearer features. Based on conventional cameras, the hardware cost is low, and the application scope is wide.DriveGaze contains original video frames of $7$ driving states(see Figure \ref{fig:sta_dataset}). The average length of videos ranges from $30$ to $464$ frames, with an average of $149.2$ frames and a standard deviation of $62.4$ frames, reflecting differences in the duration of driving states among different subjects. In total, DriveGaze includes $1645$ sequences/$245365$ frames of original events, with a total duration of $68.1$ minutes(Figure \ref{fig:sta_dataset}, divided into $1316$ for training and $329$ for testing.For more details about the dataset, such as the collection method and environment, category definition, data distribution, etc., please refer to the supplementary materials.

%% file: sections/assessment.tex
\section{Assessment}

\setlength{\tabcolsep}{3pt}
\begin{table*}[ht]
	\small
	\centering
    \resizebox{\textwidth}{!}{
	\begin{tabular}{l|c|ccccccc|cc|cc}
		\hline
		\hline 
  \multirow{2}{*}{Methods} & \multirow{2}{*}{2*} & \multicolumn{7}{c|}{Acc. of Driving Status Class (\%)} & \multicolumn{2}{c|}{Metrics (\%)} & \\
  \cline{3-13}
	 &  & Ex & Mi & An  & Br & Ti & Di & Fo & WAR $\uparrow$ & UAR $\uparrow$ & FLOPS (G) & Time (ms) \\
	\hline
Resnet18 + LSTM~\shortcite{he2016deep, hochreiter1997long} & Face & 59.1 & 80.2 & 58.8 & 53.9 & 11.3 & 81.5 & 69.1 & 59.4 & 60.8 & 7.9 & 5.0\\
Resnet50 + GRU~\shortcite{he2016deep, deng2020multitask} & Face & 28.5 & 35.4 & 45.1 & 51.6 & 8.5 & 70.0 & 6.5 & 37.1 & 36.4 & 17.3 & 10.3 \\
3D Resnet18~\shortcite{hara2018can} & Face & 56.0 & 42.3 & 61.3 & 27.6 & 45.5 & 42.7 & 91.2 & 51.8 & 52.9 & 8.3 & 21.2 \\
R(2+1)D~\shortcite{tran2018closer} & Face & 64.9 & 42.4 & 59.5 & 32.3 & 40.8 & 37.9 & 93.0 & 52.4 & 54.0 & 42.4 & 47.3 \\
Former DFER~\shortcite{zhao2021former} & Face & 83.3 & 70.1 & 77.7 & 68.9 & 45.6 & 50.7 & 90.7 & 69.4 & 70.4 & 8.3 & 7.7 \\
    \hline
Eyemotion~\shortcite{hickson2019eyemotion} & Eye & 75.9 & 79.7 & 72.0 & 86.2 & 84.6 & 79.0 & 98.5 & 83.1 & 83.3 & 5.7 & 17.5 \\
EMO~\shortcite{wu2020emo} & Eye & 76.7 & 70.0 & 63.6 & 55.8 & 45.9 & 54.1 & 95.6 & 66.6 & 66.3 & 0.3 & 7.1 \\
\hline
SEEN(E$4$-S$3$)~\shortcite{zhang2023blink} & Eye & {86.9} & 83.8 & {83.6} & 88.9 & {88.3} & {87.6} & 98.4 & {88.2} & {88.2} & 0.9 & 7.2 \\
SEEN(E$8$-S$7$) & Eye & \textbf{93.4} & \underline{90.7} & {82.6} & \underline{92.2} & \underline{93.5} & {87.8} & \textbf{99.1} & \underline{91.3} & \underline{91.3} & 0.9 & 13.4 \\
\hline
Ours(E$4$-S$3$) & Eye & \underline{91.7} & 83.8 & \underline{85.7} & 91.7 & {92.0} & \underline{94.5} & \underline{98.6} & {91.2} & {91.2} & 0.9 & 7.2 \\
Ours(E$8$-S$7$) & Eye & {90.5} & \textbf{91.2} & \textbf{90.6} & \textbf{93.2} & \textbf{94.9} & \textbf{94.9} & 98.9 & \textbf{92.4} & \textbf{92.4} & 0.9 & 13.4 \\
    \hline
	\hline
	\end{tabular}}
 \caption{ Quantitative comparison retrained and tested on the DriveGaze dataset. The abbreviations are defined as Ex $\rightarrow$ Excited; Mi $\rightarrow$ Mistake; An $\rightarrow$ Angry; Br $\rightarrow$ Brake; Ti $\rightarrow$ Tired; Di $\rightarrow$ Distracted; Fo $\rightarrow$ Focus. The first and second best results are highlighted in \textbf{bold} and \underline{underline}, respectively.}
	\label{tab:drivegaze}
\end{table*}

\setlength{\tabcolsep}{3pt}
\begin{table*}[ht]
	\small
	\centering
\resizebox{\textwidth}{!}{{
	\begin{tabular}{l|c|ccccccc|cccc|cc|cc}
		\hline
		\hline 
  \multirow{2}{*}{Methods} & \multirow{2}{*}{2*} & \multicolumn{7}{c|}{Acc. of Emotion Class (\%)} & \multicolumn{4}{c|}{Acc. under Light Conditions (\%)} & \multicolumn{2}{c|}{Metrics (\%)} & \\
  \cline{3-17}
	 &  &  Ha & Sa & An  &  Di &  Su & Fe & Ne & \makebox[0.045\textwidth][c]{Nor} & \makebox[0.045\textwidth][c]{Over} & \makebox[0.045\textwidth][c]{Low} & \makebox[0.045\textwidth][c]{HDR} & WAR $\uparrow$ & UAR $\uparrow$ & FLOPS (G) & Time (ms) \\
	\hline
     Resnet18 + LSTM~\shortcite{he2016deep, hochreiter1997long} & Face & 57.8 & 86.0 & 64.9 & 46.5 & 9.2 & 81.6 & 59.8 & 57.9 & 60.4 & 53.9 & 52.5 & 56.3 & 58.0 & 7.9 & 5.0\\
    Resnet50 + GRU~\shortcite{he2016deep, deng2020multitask} & Face & 27.9 & 38.0 & 49.7 & 44.5 & 6.9 & 70.0 & 5.6 & 43.0 & 35.7 & 28.9 & 32.8 & 35.2 & 34.7 & 17.3 & 10.3 \\
	3D Resnet18~\shortcite{hara2018can} & Face & 54.8 & 45.4 & 67.7 & 23.8 & 37.2 & 42.8 & 81.6 & 51.9 & 51.4 & 44.8 & 47.8 & 49.1 & 50.5 & 8.3 & 21.2 \\
R(2+1)D~\shortcite{tran2018closer} & Face & 63.6 & 45.5 & 65.7 & 27.8 & 33.3 & 37.9 & 86.6 & 54.3 & 50.3 & 44.4 & 49.3 & 49.7 & 51.5 & 42.4 & 47.3 \\
Former DFER~\shortcite{zhao2021former} & Face & \underline{81.5} & 75.2 & 85.8 & 59.4 & 39.3 & 50.8 & 78.6 & 70.1 & 65.4 & 66.2 & 61.1 & 65.8 & 67.2 & 8.3 & 7.7 \\
Former DFER w/o pre-train & Face & 44.1 & 65.2 &  46.0 & 66.5 & 28.0  & 50.3  & 36.1 & 47.0 & 51.9  & 45.6 & 47.2 & 48.0 & 48.0 & 8.3 & 7.7 \\
    \hline
Eyemotion~\shortcite{hickson2019eyemotion} & Eye & 74.3 & 85.5 & 79.5 & 74.3 & 69.1 & 79.2 & \underline{94.5} & 79.0 & 81.8 & \textbf{81.5} & 72.5 & 78.8 & 79.5 & 5.7 & 17.5 \\
Eyemotion w/o pre-train & Eye & 79.6 & 85.7 & 81.2 & 71.2 & 54.7 & 71.6 & \textbf{96.4} & 77.8 & 75.9 & 79.8 & 69.7 & 75.9 & 77.2 & 5.7 & 17.5 \\
EMO~\shortcite{wu2020emo} & Eye & 75.0 & 75.1 & 70.2 & 48.1 & 37.5 & 54.1 & 82.8 & 61.8 & 62.8 & 60.1 & 69.6 & 63.1 & 63.3 & 0.3 & 7.1 \\
EMO w/o pre-train & Eye & 62.0& 73.2 & 60.1 & 38.7 & 25.7 & 48.0 & 65.3 & 46.1 & 60.2 & 55.5 & 58.9 & 53.2 & 53.3 & 0.3 & 7.1\\
\hline
    SEEN(E$4$-S$3$)\shortcite{zhang2023blink} & Eye & \textbf{ 85.0} & 89.9 & \underline{92.2} & 76.7 & \underline{ 72.1} & \textbf{87.7} & 85.2 & \textbf{83.3} & 85.6 & \underline{80.8} & \underline{84.8} & \underline{83.6} & \underline{84.1} & 0.9 & 7.2 \\
    SEEN(E$7$-S$1$) & Eye & 79.0 & \underline{90.9} & 91.1 & 77.2 & 71.7 & 85.0 & 84.4 & \underline{82.4} & \underline{86.7} & 79.8 & 80.3 & 82.4 & 82.7 & 1.5 & 10.7 \\
    SEEN(E${13}$-S$0$) & Eye  & 77.9 & 88.7 & 90.2 & \underline{79.2} & 69.7 & \underline{87.6} & 84.6 & 81.1 & 86.5 & 79.4 & 81.8 & 82.3 & 82.5 & 2.6 & 19.0 \\
\hline
    Ours(E$4$-S$3$) & Eye & {78.8} & \textbf{95.0} & \textbf{97.1} & \textbf{88.3} & \textbf{ 72.1} & {75.4} & 85.0 & {81.0} & \textbf{87.6} & {80.5} & \textbf{89.0} & \textbf{84.5} & \textbf{84.5} & 0.9 & 7.2 \\
    \hline
	\hline
	\end{tabular}}}
 \caption{ Quantitative comparison retrained and tested on the SEE dataset. The abbreviations are defined as Ha $\rightarrow$ Happiness; Sa $\rightarrow$ Sadness; An $\rightarrow$ Anger; Di $\rightarrow$ Disgust; Su $\rightarrow$ Surprise; Fe $\rightarrow$ Fear; Ne $\rightarrow$ Neutrality; Nor $\rightarrow$ Normal; Over $\rightarrow$ Overexposure; Low $\rightarrow$ Low-Light. The first and second best results are highlighted in \textbf{bold} and \underline{underline}, respectively.}
	\label{tab:see}
\end{table*}

Our algorithm is not only to remember the individual's "peak" state, but also to use time clues to distinguish the states of different phases. Therefore, the main goal of this experimental part is to identify any phase of the driving state. When training and testing a state sequence, a uniformly distributed random starting point and the corresponding test time length are selected. The selection of the starting point ensures that the uniformly distributed random starting point keeps the same probability of being selected in any phase of the sequence within the closed interval from the first frame of the sequence to the sequence length minus the test time length. 
The test time length is defined by the number of event frames used $x$ and the skip time $y$ between two adjacent event frames, denoted as $E_x - S_y$. The skip time defines a window in the time domain where all events are ignored. Without loss of generality, the skip time is expressed as a multiple of $1/60$ s, that is, one frame corresponds to an event frame and a grayscale frame. $E_x - S_y$ means that the test time length is equal to $(x + (x - 1) \times y)/60$ seconds. 
Taking $E_4 - S_3$ as an example, $E_4 - S_3$ means that the number of event frames used in the end is 4; the skip time between adjacent event frames used is $3 \times 1/60$ s, that is, 3 frames are skipped; the test time length is $13/60$ s. Correspondingly, $E_8 - S_7$ means that the number of event frames used is 8 frames, 7 frames are skipped between frames, and the test time length is $57/60$ s. If the test time length cannot be met due to the short sequence length, the sequence will be read cyclically until it is met. In order to reduce the impact of randomness on the test and ensure the fairness of the comparative experiment, this paper takes randomly selected starting points for all comparative methods for testing, and takes the average of 20 tests after 20 times.We use the same random starting points for single-frame competing methods, where only the random start frame is used.
\subsection{Metrics}
To evaluate the proposed approach and compare it to competing methods, we adopt two widely used metrics: Unweighted Average Recall (UAR) and Weighted Average Recall (WAR)~\cite{2011Cross}. UAR reflects the average accuracy of different driving status classes without considering instances per class, while WAR indicates the accuracy of overall driving status; please refer to the supplementary materials for formal definitions of both metrics.
\subsection{Training Setup}
ADSN is implemented in PyTorch \cite{NEURIPS2019_9015}. We used Spike-Timing-Dependent Plasticity (STDP) for local weight adjustment and Adam optimizer with the decay rate of the first-order moment estimate set to $0.9$, the decay rate of the second-order moment estimate set to $0.999$, and the weight decay set to $1 \times 10^{-4}$ for global optimization of the network. We trained ADSN for $150$ epochs using a batch size of $128$ on an NVIDIA TITAN V GPU.For the SNN settings, we use a spiking threshold of $0.3$ and a decay factor of $0.2$ for all SNN neurons.For more details of the experiment please refer to the supplementary materials.
\subsection{Loss Function}
Because driving status recognition is a classification task, we use a regular cross-entropy loss for supervised training of ADSN:
\begin{equation}
\ell =-\frac{1}{7}\sum_{i=1}^7 y_i \log (\hat{y_i}),
\end{equation}
where  $y_i$ and $\hat{y}_i$ are the predicted $i$-th probability of driving status and corresponding ground truth probability, respectively.

\subsection{Evaluation}
We compared the effectiveness of DriveGazen with existing recognition methods (including full-face, monocular, and binocular methods) on the collected driving state dataset DriveGaze and the third-party emotion recognition dataset SEE. In order to verify the design ideas of any stage of the recognition state, three eye-based recognition methods Eyemotion\cite{hickson2019eyemotion}, EMO\cite{wu2020emo}, and SEEN\cite{zhang2023blink} were selected for comparison. In terms of network design, five common face-based temporal information methods were selected to compare with ADSN. They are ResNet18+LSTM\cite{he2016deep, hochreiter1997long}, Resnet50+GRU\cite{he2016deep, deng2020multitask}, 3D Resnet18\cite{hara2018can}, R(2+1)D\cite{tran2018closer}, and Former DFER\cite{zhao2021former}. For details on training or fine-tuning of each method, please refer to the supplementary materials. Among these previous methods, Eyemotion and EMO are single-frame methods for predicting emotions, while all other methods require full video sequences.We compare DriveGazen with SEEN during different sequence lengths.As shown in Table \ref{tab:drivegaze}, DriveGazen of E$8$-S$7$ provides the best performance of $92.4\%$ and good complexity in driving state recognition.As shown in Table \ref{tab:see}, DriveGazen of E$4$-S$3$ outperforms the runner-up method SEEN by $1\%$ in WAR and UAR on the emotion recognition dataset SEE. Our method with the same settings also outperforms SEEN by at least $4\%$ in accuracy under overexposure and HDR lighting conditions. Eyemotion performs slightly better than DriveGazen of E$4$-S$3$ in low-light conditions. We believe that Eyemotion benefits from being pre-trained on ImageNet\cite{5206848}, otherwise Eyemotion's accuracy would be $1\%$ lower than that provided by DriveGaze in the E$4$-S$3$ setting. In addition, Eyemotion requires a personalization pre-processing step, which requires subtracting an average neutral image for each person. Personalization significantly improves the accuracy of neutral emotion estimation regardless of whether Eyemotion is pre-trained on ImageNet.

\subsection{Ablation Study}
We perform a series of ablation studies on the DriveGaze dataset that investigate the impacts of input, the influence of each component of ADSN, and the impact of outputs.Table \ref{tab:ablation} summarizes the experimental results.For more detailed ablation experiments, please refer to the supplementary materials.
\setlength{\tabcolsep}{2.0pt}
\begin{table}[htbp]
	\centering
    
	\begin{tabular}{cl|cc|cc}
		\hline
		\hline
   & & \multicolumn{2}{c|}{E${4}$-S$3$} & \multicolumn{2}{c}{E${8}$-S$7$} \\
   \hline
    & Networks    &WAR &UAR    &WAR &UAR   \\
    \hline
    $A$	& w/o $I^n$  & 88.6 & 88.7 & 89.7 & 89.8\\  
    $B$	& $I^n \rightarrow$ $I^2$    & 89.2 & 89.2 & 90.3 & 90.3\\  
    $C$ & $[I^1,...,I^n]$   & 90.4 & 90.4 & 91.5 & 91.5\\ 
    \hline
    \hline
    $D$	& No Multi-head self-attention & 89.5 & 89.6 & 90.7 & 90.9\\ 
    $E$ & No $F_s \rightarrow J_s$ & 90.5 & 90.3 & 91.6 & 91.4\\
    $F$	& No Residual  & 84.0 & 83.8 & 85.1 & 84.9\\
    $G$	& No guide attention  & 65.4 & 65.0 & 66.1 & 65.8\\
    \hline
    $H$ & SNN $\rightarrow$ CNN    & 60.7 & 60.3 & 61.4 & 61.0\\ 
    $I$ & SNN $\rightarrow$ LSTM   & 60.8 & 60.4 & 61.5 & 61.1\\ 
    \hline
    \hline
    $J$	& Last potential   & 88.5 & 88.6  & 89.6 & 89.8 \\
    $K$	& Last spike   & 68.9 & 68.2  & 69.8 & 69.1\\ 
    \hline
    \hline
    $M$	& Ours  & 91.2 & 91.2 & 92.4 & 92.4\\ 
	\hline
	\hline
\end{tabular}

 \caption{Ablation comparisons show that: both the first and last intensity frames are essential for providing discriminative features; all components of ADSN contribute to the overall performance ; and potential averaging is necessary results in a more accurate performance.}
    \label{tab:ablation}
\end{table}

\textit{Impacts of Input.} ADSN leverages the first and last intensity frames. Experiments (A), (B) and (C) gauge the impact of the intensity frames: experiment (A) only uses the first intensity frame, experiment (B) replaces the last intensity frame with the second frame, and experiment (C) uses all the intensity frames corresponding to the included event frames.  

\textit{Influence of ADSN components.} We investigate the effectiveness of the different components that comprise ADSN: 1) the effectiveness of the attention, residual and spatiotemporal features fusion(experiments (D) and (G)) and 2) the benefits of SNNs (experiments (H) to (I)).  

\textit{Impact of outputs.}
ADSN estimates status based on the average of $n$ membrane potentials; see Equation \ref{eq:T_O} and Equation \ref{eq:final_result}. Instead of using the average of $n$ membrane potentials, we define the prediction score based on the potential generated by the last event frame only (experiment (J)); similar to the previous but using output spikes instead of potential (experiment (K)). 

%% file: sections/conclution.tex
\section{Conclusion}
We introduce a novel wearable prototype for driver status recognition, which can effectively estimate the driver's status under challenging lighting conditions. We investigated recognition based on input from conventional cameras. Conventional cameras have high resolution, captured frames can robustly encode spatial information. However, parsing temporal cues is challenging. We introduce DriveGazen, a learning-based novel solution for extracting informative temporal cues for status recognition. DriveGazen introduces several novel design components: a method for generating DVS event frames from video frames,a spatial feature extractor based on multi-scale self-attention,a CNN-SNN-based temporal feature extractor and guide attention mechanism. Leveraging spatial awareness and the pulse mechanism of SNN to effectively provide discriminative features for classification. Spatial attention is injected into temporal feature extraction during both training and inference stages. Our extensive experimental results demonstrate that DriveGazen can effectively estimate driver status at any stage. To the best of our knowledge, DriveGazen is the first attempt to utilize conventional camera-generated events and guided attention SNN for driving status recognition tasks.
